%% file: emnlp2023.tex
\pdfoutput=1

\documentclass[11pt]{article}

\usepackage{amsmath}
\usepackage{EMNLP2023}
\usepackage{hyperref}

\usepackage{times}
\usepackage{latexsym}
\usepackage{graphicx} 
\usepackage{subfig}
\usepackage[T1]{fontenc}

\usepackage[utf8]{inputenc}

\usepackage{microtype}

\usepackage{inconsolata}

\title{Non-contrastive sentence representations via self-supervision}

\author{Marco Farina\thanks{\enspace Alphabetical order} \\
  Bloomberg \\
  \texttt{mfarina19@bloomberg.net} \\\And
  Duccio Pappadopulo\footnotemark[1] \\
  Bloomberg \\
  \texttt{dpappadopulo@bloomberg.net} \\}

\begin{document}
\maketitle
\begin{abstract}

Sample contrastive methods, typically referred to simply as \emph{contrastive} are the foundation of most unsupervised methods to learn text and sentence embeddings. On the other hand, a different class of self-supervised loss functions and methods have been considered in the computer vision community and referred to as \emph{dimension contrastive}. In this paper, we thoroughly compare this class of methods with the standard baseline for contrastive sentence embeddings, SimCSE~\citep{simcse}. We find that self-supervised embeddings trained using dimension contrastive objectives can outperform SimCSE on downstream tasks without needing auxiliary loss functions.
\end{abstract}

\section{Introduction}
Text embeddings are an important tool for a variety of NLP tasks. They provide a general and compute efficient solution to problems like topic classification, document clustering, text mining and information retrieval, among others.

Most modern techniques to learn text embeddings rely on minimizing a contrastive loss~\cite{contr, infonce}. This requires identifying, for each example $x$ in the training set, a \emph{positive} example $x^+$ and a set of \emph{negative} examples $x^-_i$ associated to $x$.
The choice of $x^+$ and $x_i^-$ is one of the main factors differentiating these techniques. Unsupervised methods~\cite{emmim, declutr, diffcse} rely on in-batch negatives for the $x_i^-$ and data augmentation for $x^+$. Supervised or weakly supervised methods~\cite{sentencebert, gtr, e5, instructor, sgpt, st5} rely either on mining heuristics or annotated datasets to build the positive and negative pairs, for instance a  common choice is to use entailment and contradiction pairs respectively, as in  SNLI~\cite{bowman2015large} and MNLI~\cite{williams2018broadcoverage}.

In this work we approach the problem of learning text embedding from the point of view of which objective function to use. We consider two self-supervised representation learning algorithms introduced in the computer vision literature: Barlow Twins (BT)~\cite{barlow} and VICReg~\cite{vicreg}.

What teases apart these two methods is their nature of being \emph{dimension contrastive} according to the classification of \citet{ssl2}: while the usual contrastive method, defined by \citet{ssl2} as \emph{sample contrastive}, avoids the collapse of the learned representations by penalizing similarity of the embeddings corresponding to different data points, dimension contrastive methods regularize the objective function by de-correlating the embeddings across their dimensions. Both sample and dimension contrastive methods rely on data augmentation in the unsupervised setting. While good augmentation functions are known and routinely used for image data, augmentation of textual data is usually considered trickier~\citep{dasurvey}. One of the breakthrough of SimCSE is the realization that using the model stochastic dropout mask to define the augmented views of the same data point is an effective choice.

The main goal of this paper is to compare the embeddings learned through sample-contrastive and dimension-contrastive techniques and explore different augmentation strategies. We use SimCSE~\cite{simcse} as our sample-contrastive baseline and compare it against BT and VICReg\footnote{To the best of our knowledge we are first to use VICReg as an objective to train sentence embeddings.}. Our main findings are: i) Barlow Twins is competitive with unsupervised SimCSE as a standalone objective function and outperforms it on a majority of MTEB tasks with a RoBERTa based architectures. This is partly at odds with the finding of \citet{scoop1} and \citet{scoop2} which include new terms in the loss with the motivation that BT alone does not get better performances than SimCSE. A thorough comparison of dimension and sample contrastive methods does not exist in the literature. ii) VICReg underperforms Barlow Twins and SimCSE: we find it harder to optimize it and we cannot exclude that more hyperparameter exploration and better data augmentation would lead to better results. iii) We obtain mixed results by using supervision (for instance from NLI datasets) in place of data augmentation: in no case supervision leads to better performances across all MTEB downstream task categories.

\section{Contrastive techniques}

\begin{table*}[ht]
\centering
\input{table_aug}
\caption{\label{tab:aug} max, upper quartile, and median for the distribution of STS-B Spearman's correlations on the dev set as a function of the data augmentation. Bold: overall best score per model, underlined: best score per augmentation. For VICReg we only ran EDA with with $\alpha=0.1$.}
\end{table*}

All the techniques that we experiment with in the following can be described in a unified way.
Consider a batch of data points $s_n$, $n=1,\ldots,N$ (sentences in this work).\footnote{We use $n,m$ to denote different members of the same batch and $i,j,k$ to denote different  dimensions in the same embedding.}
The representation $\bf{e}_n$ for each point is obtained through a parametrized sentence encoder (BERT and RoBERTa are what we will use in this paper): ${\bf{e}}_n = E_\theta(s_n)$. In order to consider data augmentation of any type, we assume that $E_\theta$ allows for a second (possibly random) parameter $\bf{\epsilon}$ specifying the augmentation ${\bf{e}}'_n = E_\theta(s_n, \bf{\epsilon})$. When training $E_\theta$ in the self-supervised setting we create two embeddings (\emph{views}) of each point in the batch, ${\bf{e}}^{(A,B)}_n$. Each of them is projected to a high-dimensional space by means of a parametrized \emph{projector} ${\bf{z}}_n \equiv P_\theta({\bf{e}}_n)$. The resulting D-dimensional vectors ${\bf{z}}_n$ are then used in the method specific loss function.

{\bf{SimCSE}} $-$ Our baseline for sample contrastive methods is SimCSE~\cite{simcse}. According to the previous definitions the unsupervised version of SimCSE minimizes the contrastive loss
\begin{equation}\label{simcse}
    \Delta L_{\textrm{SimCSE}} = -\log \frac{e^{{\textrm{sim}}({\bf{z}}^{(A)}_n, {\bf{z}}^{(B)}_n)/\tau}}{\sum_{m}e^{{\textrm{sim}}({\bf{z}}^{(A)}_n, {\bf{z}}^{(B)}_m)/\tau}}
\end{equation}
summed over the batch $n=1,\ldots, N$. ${\textrm{sim}}$ is a similarity function, in this case the standard cosine similarity. Unsupervised SimCSE uses different dropout masks applied to the same input data point to obtain the two views of the same sample.

{\bf{Barlow Twins}} $-$ BT~\cite{barlow} is one of the two dimension contrastive methods we consider. Each batch contributes to the loss by an amount
\begin{equation}\label{barlow}
\Delta L_{BT} = \sum_i(1-\rho_{ii})^2+\lambda_{\textrm{BT}} \sum_{j\neq i} \rho_{ij}^2
\end{equation}
where $\rho_{ij}$ is the Pearson correlation between the $i$-th and $j$-th entry of the embeddings of ${\bf{z}}^{(A)}$ and ${\bf{z}}^{(B)}$. The first term in Eq.~\ref{barlow} enforces that the embedding of the two views A and B are perfectly correlated; the second term on the other hand regularizes the first and requires different embedding components to be uncorrelated and ideally to encode different information about the data.

{\bf{VICReg}} $-$ The second example of dimension contrastive technique that we examine is VICReg~\cite{vicreg}. In this case the loss function combines three terms 
\begin{align}\label{vicreg}
    &L_{\textrm{VICReg}}  = \frac{\lambda_I}{N}\sum_n ||{\bf{z}}_n^{(A)} - {\bf{z}}_n^{(B)}||^2 +\\\nonumber
    & \frac{\lambda_V}{D}\sum_{i, I} H\left(\sqrt{C^{(I)}_{ii}+\epsilon}\right)  + \frac{\lambda_C}{D} \sum_{i\neq j, I} C^{(I)}_{ij}{}^2
\end{align}
where $I=A, B$, and $H = \max(0, 1-x)$. The $D\times D$ matrix $C$ in Eq.~\ref{vicreg} is the covariance matrix for the component of the ${\bf{z}}^{(A, B)}$ vectors estimated within a batch. Similarly to BT the first term in the loss drives two views of the same data point to be represented by the same vector while the other two terms are introduced to prevent embeddings' collapse. The last term in Eq.~\ref{vicreg} has similarities with the regularization criteria used by BT, and it tries to de-correlate different components of the vectors ${\bf{z}}^{(A, B)}$; the second term is a hinge loss that encourages the variance of each of the components of the same vectors to be of order 1. 

There is extensive work trying to understand the representation learned by contrastive (\citet{isolawang} \emph{inter alia}) and non-contrastive methods (\citet{ssl1, ssl2, ssl3} \emph{inter alia}) and the reason of their success. Among these works we wish to point out~\citet{ssl2} in which the similarities between sample-contrastive and dimension-contrastive objectives are extensively discussed and the different performances of the two classes of methods, albeit in the vision domain, are attributed to architectural and hyperparameter choices. Ultimately which of these methods work better in the text modality is an empirical question and attempting to answer this question is the main goal of this paper.

\section{Methods}
\label{sec:methods}

In order to compare with \citet{simcse}, we use the same Wikipedia dataset\footnote{The dataset can be downloaded at this \href{https://huggingface.co/datasets/princeton-nlp/datasets-for-simcse/resolve/main/wiki1m_for_simcse.txt}{link}.}  they used to train the unsupervised models.

For our supervised experiments we try two datasets. The first, used also by \citet{simcse}, is the set of entailment pairs from SNLI~\cite{snli} and MNLI~\cite{mnli}. Only the positive pairs are used, as hard negatives cannot be easily incorporated in our objectives. The other is WikiAuto~\cite{wikiauto}, a set of sentences from English Wikipedia aligned to their simplified English Wikipedia version.

We consider two base models for our experiments, BERT-base and RoBERTa-base. In each case the embedding $E_\theta$ that we use for downstream tasks is the embedding of the \texttt{[CLS]} token. The projector $P_\theta$ for SimCSE is a linear layer with the same dimension as the transformer dimension, followed by $\tanh$ activation. For BT and VICReg we follow \citet{vicreg} and use two linear layers with batch normalization and ReLU activation, followed by an additional linear layer all of dimension 8192. Larger dimensions give similar results and smaller ones progressively degrade performances.

The SimCSE models are trained with a temperature $\tau=0.05$, and a learning rate of $3\times10^{-5}$ for BERT and $10^{-5}$ for RoBERTa, which were identified with an hyperparameter sweep.

We experiment with three basic types of augmentations for BT and VICReg. \underline{Dropout}: as in \citet{simcse} we apply different dropout masks to each view of the same data point; this augmentation is parametrized by the dropout probability $p_{\textrm{do}}=\{0.05, 0.1, 0.2\}$. \underline{Shuffling}: for both branches we select a fraction $p_{\textrm{s}} = \{0.05, 0.1, 0.2, 0.3, 0.5\}$ of the input tokens and apply a random permutation. \underline{EDA}~\cite{eda}: we apply EDA to each branch with the same parameter $\alpha=\{0.1, 0.2\}$ for synonym replacement, random insertions, random swaps, and random deletions. For each augmentation we perform a hyperparameter scan to select the best value of the remaining parameters (learning rate and the loss coefficients in Eqs.~\ref{barlow} and \ref{vicreg}). We measure the Spearman's rank correlation on the STS-B~\cite{Cer_2017} validation set to select the best checkpoints as in \citet{simcse}. 

Results are shown in Tab.~\ref{tab:aug}. Across models and loss functions, smaller $p_{\textrm{do}}$ and larger $p_{\textrm{shuffle}}$ values are preferred, and the effect is more pronounced with BT. EDA underperforms in all cases. For more details about the scans we refer to Appendix \ref{sec:hyperparameters}.

\section{Results}

We evaluate the embedding on a variety of downstream tasks using the Massive Text Embedding Benchmark (MTEB)~\cite{mteb} and report both average performances on the test set and a breakdown by task category in Table~\ref{tab:mteb}\footnote{We refer to Appendix~\ref{sec:mteb} for a summary of the task contained in the benchmark and a complete breakdown of the scores by task.}.

\begin{table*}[ht]
\centering
\input{mteb_table}
\caption{MTEB test performances aggregated by task category for (Ro)BERT(a): average of last layers, SimCSE{\footnotemark} and our best hypertuned models from Tab.~\ref{tab:aug}. We display the performances of the best models for both dropout and shuffle augmentations with overall best scores in bold. We also include results from best RoBERTa Barlow Twins models trained on alternative datasets underlying best scores. Alignment and uniformity are also shown.}
\label{tab:mteb}
\end{table*}

While BERT scores trail behind SimCSE by a few percent points for both  BT and VICReg for the majority of tasks, RoBERTa with BT and dropout outperforms SimCSE with two notable exceptions: pair classification and STS. For pair classification we notice that embeddings trained using shuffle augmentation outperform those trained with dropout irrespectively of model architecture or objective. The STS results seem to indicate some degree of overfitting to the STS-B dev set. This seems more severe for VICReg for which the dev set performances in Tab.~\ref{tab:aug} are above BT. 

Evaluating on STS tasks is a common practice which we also follow to select model checkpoints. However, this has been criticized due to the lack of correlation between STS performances and downstream task performances~\cite{reimers-etal-2016-task, wang-etal-2021-tsdae-using, abe-etal-2022-sentence}. Finally we notice that models trained on supervised datasets can outperform unsupervised methods on certain downstream tasks, but there is no clear winner. This aligns with the finding of~\citet{mteb} in which single model performance on different tasks
varies a lot with no single model winning across all tasks.

We also report \emph{alignment} and \emph{uniformity}, two metrics which are commonly considered when analyzing sample contrastive embedding techniques: the standard sample contrastive objective optimizes them in the limit of infinitely many negative samples~\citep{isolawang}. They are shown to empirically correlate to the embedding performances on downstream tasks, but an understanding of why uniformity is needed is lacking. \citet{huang2023generalization} derives an upper bound on the error rate for classification tasks based on three metrics, alignment, \emph{divergence}, and \emph{concentration}. Intuitively, the latter two represent how separated the centroids of the various classes are in the embedding space and how concentrated around such centroid are the representation of the augmented members of each class. \citet{huang2023generalization} show that both the InfoNCE~\cite{infonce} and BT satisfy these criteria. We refer to Appendix~\ref{sec:aluf} for further discussions of alignment an uniformity. 

\footnotetext{SimCSE scores differ from those reported in \citet{mteb} because we evaluate unsupervised models without projector consistently with what done in \citet{simcse}.}

\section{Conclusions}

In this work, we compare sample contrastive (SimCSE) and dimension contrastive (Barlow Twins, VICReg) training objectives to learn sentence embeddings. Our results shows how these  alternative self-supervision objectives can learn good representations, performing as well as or better than those obtained from SimCSE. Dimension contrastive techniques are largely unexplored outside the computer vision literature and we hope this work could be a step in the direction of popularizing them in the NLP community.

\section*{Limitations}
The goal of this short paper is to make the point that dimension contrastive objectives are a viable alternative to standard sample contrastive techiniques. 

While we used SimCSE as our baseline, it would be interesting to use sample contrastive loss functions on methods like DiffCSE~\cite{diffcse}, InfoCSE~\cite{infocse} and PromptBERT~\cite{jiang-etal-2022-promptbert} and see whether the same improvement in performance obtained using the standard contrastive loss function would apply to BT or VICReg.

It would be interesting to study different model architectures like decoder-only models~\cite{sgpt} or encoder-decoder ones~\cite{st5}.

Additionally, while our study is limited to sentence embeddings for English documents, the methods are applicable to multilingual corpora and it would be worth exploring them in this context.

\bibliography{anthology,custom}
\bibliographystyle{acl_natbib}

\appendix

\section{Hyperparameters}
\label{sec:hyperparameters}

In the hyperparameter search the model architectures are fixed both in terms of the base models (BERT and RoBERTa) and in terms of the projectors that are used (see Sec.~\ref{sec:methods}). We furthermore fix the batch size to 256 as we did not observe significant gains with larger batches.

All models are trained for 2 epochs. We evaluate every 60 steps and the final metric we use for checkpoint selection is the Spearman's correlation on the STS-B dev set.

\subsection{Barlow Twins}
For BT we use a grid scan to explore hyperparameters and data augmentations. We use the values reported in Table~\ref{table:ht} for both BERT and RoBERTa models. Augmentations are not combined, but for each augmentation we scan learning rate and the loss coefficient ($\lambda_{\textrm{BT}}$).

\renewcommand{\arraystretch}{1.2}
\begin{table}
\small
\centering
\begin{tabular}{ll}
\hline
Parameter & Domain\\
\hline
learning rate& $\{1, 2, 5\}\times 10^{-5}$\\ 
dropout& $\{0.05, 0.1, 0.2\}$ \\
shuffle& $\{0.5, 1, 2, 3, 5\}\times 10^{-1}$ \\
EDA& $\{0.1, 0.2\}$ \\
\hline
\multicolumn{2}{c}{Barlow Twins} \\
\hline
$\lambda_{\textrm{BT}}$& $\{0.5, 1, 2.5, 5, 7.5, 10, 25\}\times 10^{-3}$ \\ 
\hline
\multicolumn{2}{c}{Barlow Twins} \\
\hline
$\log_{10}\lambda_{\textrm{C}}$& $[-3, -1]$ \\ 
$\log_{10}\lambda_{\textrm{V}}$& $[1, 4]$ \\ 
shuffle& $\{0.5, 1, 2, 3, 5\}\times 10^{-1}$ \\
\hline
\end{tabular}
\caption{\label{table:ht} BERT and RoBERTa values for the BT and VICReg hyperparameter scan. The scan over $\lambda_{\textrm{C, V}}$ is uniform in log space. For VICReg we only use $\alpha=0.1$ for EDA augmentation.}
\end{table}

\begin{figure}[t]
    \includegraphics[scale=0.5]{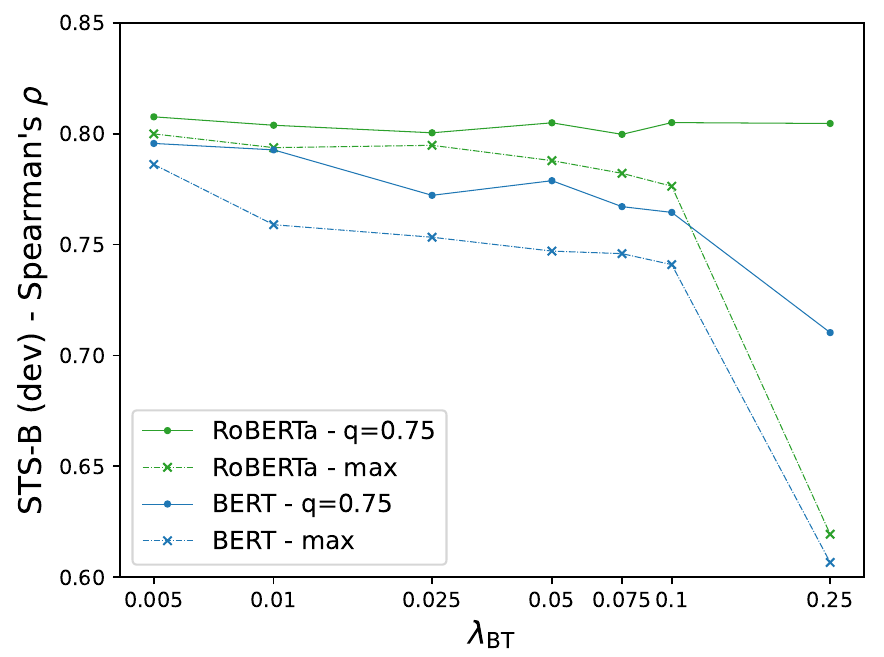}
  \caption{\label{fig:lambda_bt}  STS-B performances as a function of the $\lambda_{\textrm{BT}}$ coefficient. We show both the max and the upper quartile of the metric distribution after binning by the value of the parameter.
  }
\end{figure}

We find the performances to be quite insensitive to the choice of the learning rate, but quite sensitive to $\lambda_{\textrm{BT}}$ for both model architectures. This is shown in Fig.~\ref{fig:lambda_bt}. We thus constrain $\lambda_{\textrm{BT}}\leq 0.05$ for BERT and $\leq 0.025$ for RoBERTa. We show the development set performances as a function of the augmentation in Tab.~\ref{tab:aug}.

\begin{figure}[t]
    \includegraphics[scale=0.5]{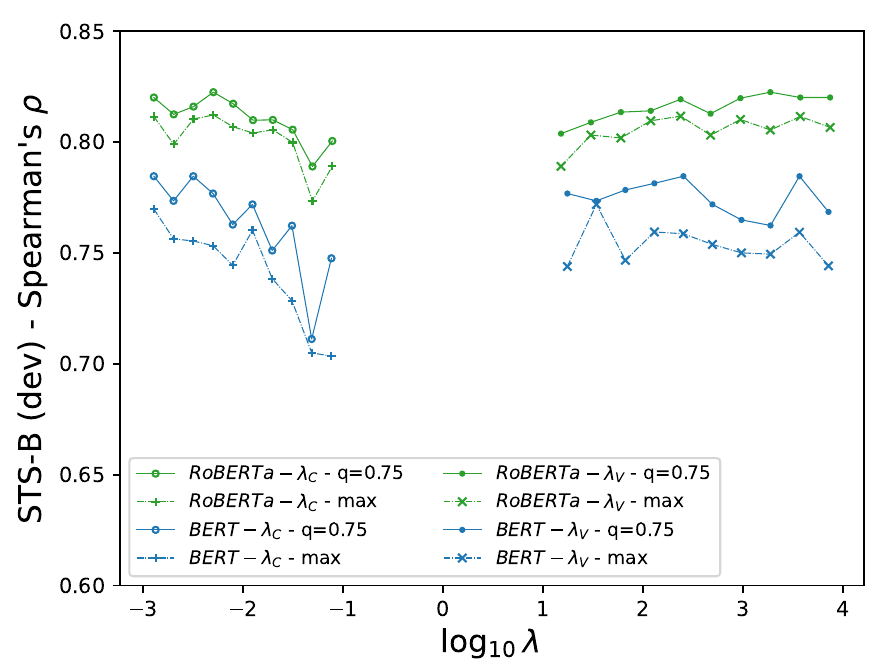}
  \caption{\label{fig:lambda_vic}  STS-B performances as a function of $\lambda_{\textrm{C}}$ and $\lambda_{\textrm{V}}$. We show both the max and the upper quartile of the metric distribution after binning by the value of the parameter.
  }
\end{figure}

\subsection{VICReg}
The parameter space of VICReg is larger than the one of BT: the loss function depends on 3 parameters $\lambda_{\textrm{V}, \textrm{I}, \textrm{C}}$. We fix $\lambda_{ \textrm{I}}=1$ and scan the remaining two parameters. Since the parameter is larger we use SMAC instead of grid search.
Table~\ref{table:ht} report the parameters of the scan. Similarly to BT augmentations are not combined, but for each augmentation we scan learning rate, $\lambda_{\textrm{V}}$, and $\lambda_{\textrm{C}}$. For each augmentation strategy we run a total of 50 jobs.

Similarly to BT, there is little sensitivity to the learning rate. We find that the scan favors small values of $\lambda_{ \textrm{C}}$ and large values of $\lambda_{ \textrm{V}}$. The dev set performances as a function of the augmentation are shown in Tab.~\ref{tab:aug}.

\section{Alignment and uniformity}
\label{sec:aluf}

\begin{figure}[t]
    \includegraphics[scale=0.5]{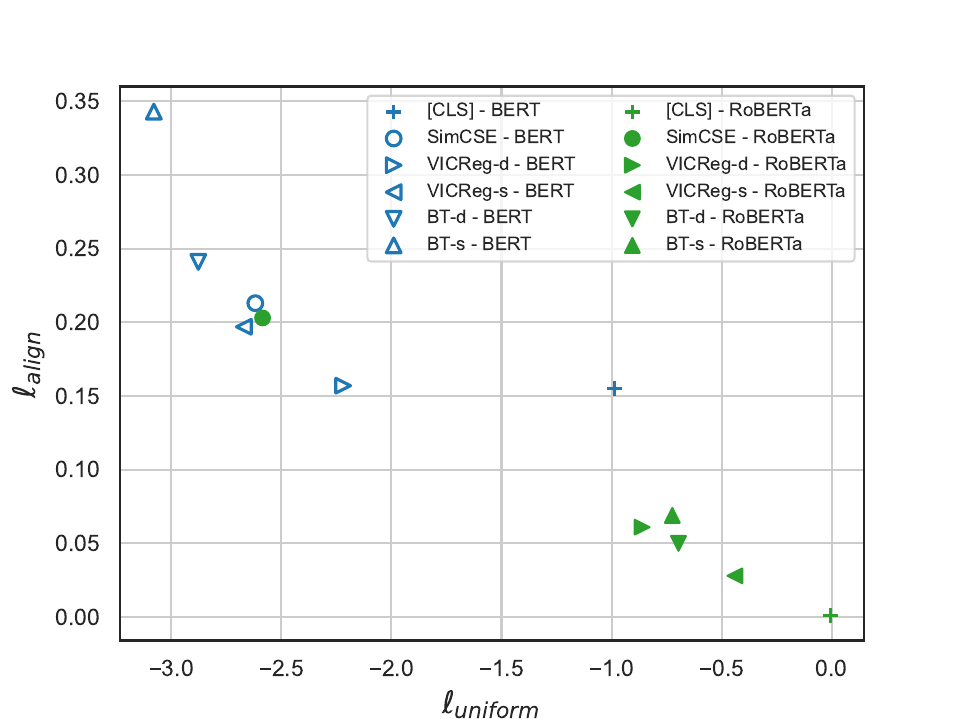}
  \caption{\label{fig:aluf}  Alignment and uniformity numbers for the models reported in Tab.~\ref{tab:mteb}. \texttt{[CLS]}-(Ro)BERT(a) represent text embedding models obtained by using the last layer \texttt{[CLS]} token as the embedding. Lower values are better for both metrics.
  }
\end{figure}

We calculate the alignment and uniformity metrics~\cite{isolawang} for the unsupervised models shown in Tab.~\ref{tab:mteb}. Optimizing the unsupervised objective, either sample or dimension contrastive, improve uniformity in all cases while it typically degrades alignment. We notice that these effects are particularly pronounced for the sample contrastive objective optimized by SimCSE, in particular in terms of the improvement in uniformity.

\input{histos}

For both BT and VICReg, and in particular for RoBERTa, uniformity improves only marginally through training. However this does not seem to hurt performances on downstream tasks as shown in Tab.~\ref{tab:mteb}. This is consistent with the discussion of \citet{huang2023generalization}. 

Another representation of this fact is Fig.~\ref{fig:histos} which shows the distribution of cosine similarities of sentence pairs on the STS-B test set stratified by the similarity rating assigned by human annotators. We see that both SimCSE, BT, and VICReg training increase the divergence of the distributions across buckets, but SimCSE tends, on average, to achieve that by spreading the embeddings apart on the hypersphere (notice the different horizontal scale of the 3 bottom panels in Fig.~\ref{fig:histos})

\section{MTEB}
\label{sec:mteb}
The MTEB (Massive Text Embedding Benchmark)~\cite{mteb} is a comprehensive evaluation tool designed to assess the performance of text embedding models. It includes well established benchmarks, and spans a wide range of tasks and domains. 

We report results on the 56 English language datasets. They are divided in the following tasks (associated evaluation metrics in parenthesis): Classification (accuracy), Clustering (v-measure), Pair Classification
(average precision), Rerank (MAP), Retrieval (nDCG@10), STS (Spearman correlation), and Summarization (Spearman correlation). 
A breadkdown of all datasets, compiled with results from our RoBERTa models, is shown in Tab.~\ref{tab:bigmteb}.

\begin{table*}[t]
\centering
\setlength\tabcolsep{4pt}
\input{bigtable_mteb}

\caption{MTEB performances of RoBERTa models on all English datasets grouped by task. We display the scores for both dropout and shuffle augmentations with overall best scores in bold. We also include scores from best Barlow Twins models trained on alternative datasets underlying best scores. 
$\diamondsuit$: custom clustering datasets created for MTEB, for details we refer to \citet{mteb}. $\spadesuit$: retrieval datasets are a subset of the BEIR benchmark~\cite{beir}. $\heartsuit$: tasks from the original STS benchmark~\cite{agirre2012semeval, agirre2013sem}.
}
\label{tab:bigmteb}
\end{table*}

\end{document}

%% file: table_aug.tex
\small
\renewcommand{\arraystretch}{1.25}
\begin{tabular}{ll| c c c |c c |c c c c c}
    \hline
     && \multicolumn{3}{c|}{dropout ($p_{\textrm{do}}$)} & \multicolumn{2}{c|}{EDA ($\alpha$)} & \multicolumn{5}{c}{shuffle ($p_{\textrm{shuffle}}$)}\\
       && 0.05    & 0.1    & 0.2    & 0.1       & 0.2    & 0.05     & 0.1    & 0.2 & 0.3    & 0.5   \\ \hline\hline  \multicolumn{12}{c}{\textbf{Barlow Twins}} \\\hline 
    BERT&max & \underline{77.9} & 74.0 & 73.5 & \underline{74.3} & 73.9 & 76.6 & 77.8 & 78.9 & 79.5 & \textbf{79.6} \\ \hline
    & q75 & 75.1 & 73.2 & 72.4 & 72.9 & 72.4 & 75.0 & 76.7 & 78.0 & 78.8 & 78.6 \\ \hline
    & q50 & 74.0 & 72.6 & 72.2 & 72.5 & 71.6 & 73.7 & 75.8 & 76.0 & 77.6 & 77.7 \\ \hline
    RoBERTa&max & 80.0 & \textbf{80.5} & 78.1 & 76.0 & \underline{77.2} & 79.5 & 80.4 & 80.2 & 80.4 & \underline{80.8} \\ \hline
    & q75 & 78.6 & 77.4 & 77.0 & 74.2 & 75.8 & 78.2 & 80.0 & 79.9 & 80.1 & 80.0 \\ \hline
    & q50 & 78.0 & 75.2 & 74.4 & 73.1 & 74.4 & 77.6 & 78.7  & 79.4 & 79.8  & 79.5 \\ \hline\hline \multicolumn{12}{c}{\textbf{VICReg}} \\\hline 
        BERT&max & \underline{76.2} & 75.3 & 75.5 & 76.0 & \underline{76.3} & 77.6 & 76.8 & 77.4 & 78.1 & \textbf{78.5} \\ \hline
    & q75 & 74.8 & 74.2 & 74.0 & 75.0 & 75.1 & 76.4 & 75.4  & 77.2 & 77.8 & 77.7 \\ \hline
    & q50 & 74.5 & 73.5 & 73.0 & 74.2 & 74.2 & 75.3 & 73.8 & 77.0 & 75.9 & 77.2 \\ \hline
    RoBERTa&max & 81.2 & 81.0  & \underline{81.6} & 80.2 & \underline{80.4} & 82.0 & 81.9  & 81.6 & 82.2 & \textbf{82.0} \\ \hline
    & q75 & 80.7 & 80.4 & 80.3 & 79.0 & 79.3 & 79.7 & 80.9 & 81.3 & 81.3 & 81.8 \\ \hline
    & q50 & 80.4 & 80.0 & 79.7 & 78.0 & 77.3 & 79.0 & 80.0  & 81.2 & 81.0  & 81.3 \\ \hline  
\end{tabular}

%% file: mteb_table.tex
\small
\renewcommand{\arraystretch}{1.25}
\setlength{\tabcolsep}{5.5pt}
\begin{tabular}{l c c c c c c c c | c c }
    \hline
    \textbf{Method} & \textbf{Class.} & \textbf{Clust.} & \textbf{PairClass.} & \textbf{Rerank.} & \textbf{Retr.} & \textbf{STS} & \textbf{Summ.} & \textbf{Avg.} & $\ell_{\textrm{align}}$ & $\ell_{\textrm{unif}}$ \\  
    \hline
    \multicolumn{11}{c}{BERT} \\ \hline
    avg. & 61.7 & 30.1 & 56.3 & 43.4 & 10.6 & 54.4 & 29.8 & 38.3 & 0.20 & -1.62\\
    SimCSE & 63.7 & 30.5 & 73.1 & \textbf{47.0} & \textbf{21.5} & \textbf{74.8} & \textbf{31.2} & \textbf{46.6} & 0.21 & -2.62\\ 
    VICReg ($p_{\textrm{do}}=0.05$) & 62.9 & 33.0 & 61.8 & 46.0 & 17.4 & 67.8 & 29.3 & 43.9 &
    0.16 & -2.22 \\ 
    VICReg ($p_{\textrm{shuffle}}=0.5$) & 59.0 & \textbf{33.3} & 63.8 & 46.1 & 19.3 & 67.7 & 29.8 & 43.7 & 0.20 & -2.67 \\ 
    Barlow Twins ($p_{\textrm{do}}=0.05$) & \textbf{63.7} & 29.9 & 69.4 & 46.3 & 18.7 & 70.0 & 30.1 & 44.6 & 0.24 &  -2.88\\ 
    Barlow Twins ($p_{\textrm{shuffle}}=0.5$) & 59.1 & 27.9 & \textbf{73.4} & 45.7 & 16.6 & 70.6 & 29.0 & 42.9 & 0.34 & -3.08\\ 
    \hline
    
    \multicolumn{11}{c}{RoBERTa} \\ \hline
    avg. & 60.0 & 21.6 & 54.1 & 40.2 & 5.8 & 53.8 & 29.6 & 34.6 & 0.01 & -0.16\\ 
    SimCSE & 64.6 & 30.8 & \textbf{74.5} & \textbf{47.3} & 23.6 & \textbf{74.4} & 27.7 & 47.4 & 0.20 & -2.59\\ 
    VICReg ($p_{\textrm{do}}=0.2$) & 61.3 & 33.4 & 68.2 & 46.1 & 19.9 & 70.5 & 28.7 & 45.1 & 0.06 & -0.86 \\ 
    VICReg ($p_{\textrm{shuffle}}=0.5$) & 63.0 & 32.4 & 70.7 & 47.3 & 20.7 & 70.6 & \textbf{29.2} & 45.7 & 0.03 &  -0.44\\ 
    Barlow Twins ($p_{\textrm{do}}=0.1$) & \textbf{65.2} & \textbf{33.9} & 70.6 & \textbf{47.3} & \textbf{24.1} & 71.1 & 28.9 & \textbf{47.5}&  0.05 & -0.70 \\ 
    Barlow Twins ($p_{\textrm{shuffle}}=0.5$) & 59.4 & 28.1 & 73.1 & 45.3 & 21.5 & 72.2 & 27.6 & 44.5 & 0.07 & -0.72 \\
    \hline

    Barlow Twins (NLI) & \underline{60.3} & \underline{36.8} & \underline{71.2} & \underline{47.6} & 25.1 & 70.0 & 27.5 & \underline{47.1} & 0.01 & -0.12 \\
    Barlow Twins (WikiAuto) & 58.1 & 33.5 & 67.7 & 45.6 & \underline{25.9} & \underline{70.6} & \underline{31.1} & 46.0 & 0.01 & -0.11 \\
    \hline
\end{tabular}

%% file: histos.tex
\begin{figure}[t]%
    
    \centering
    \subfloat[RoBERTa \texttt{[CLS]}]{\includegraphics[scale=0.44]{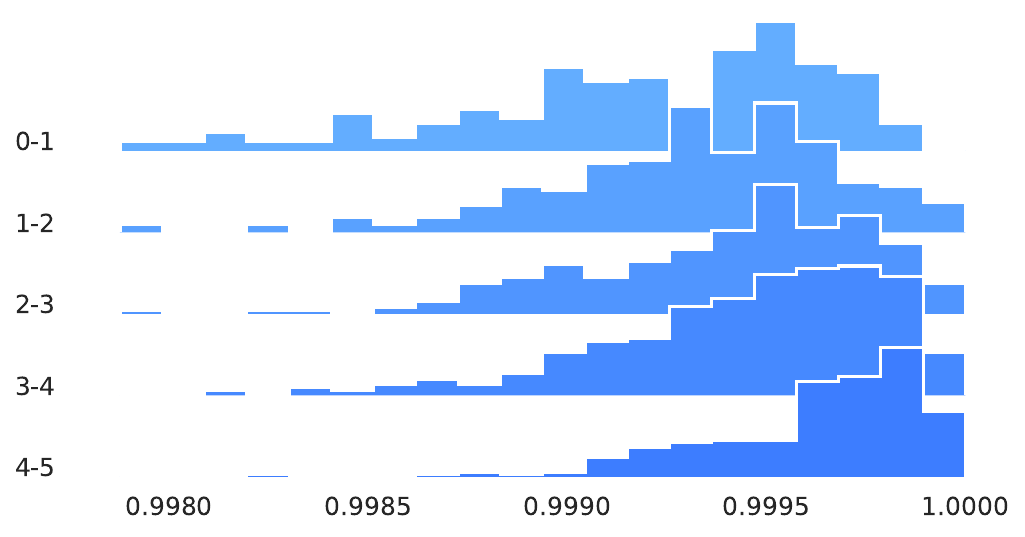}}

    \subfloat[SimCSE RoBERTa]
    {\includegraphics[scale=0.44]{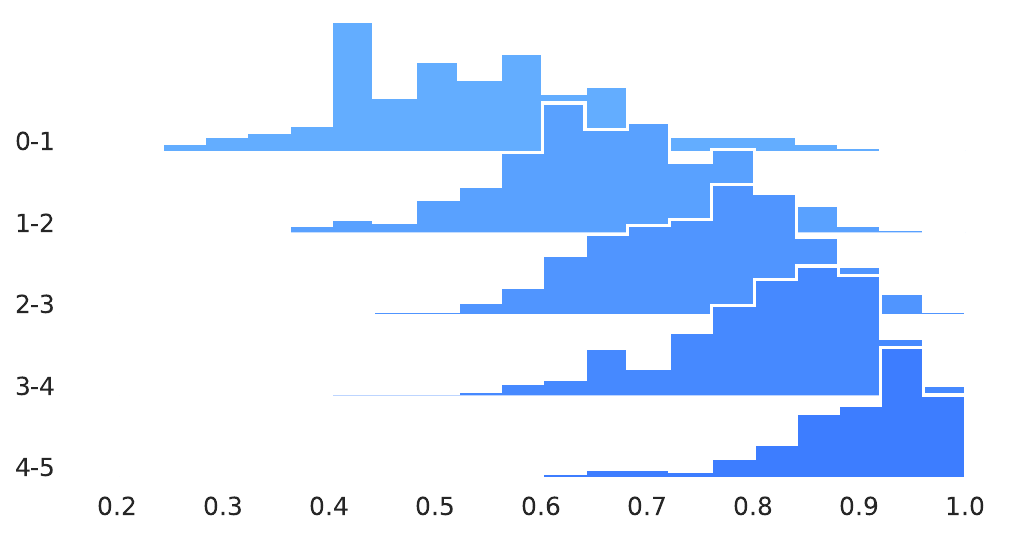}}
    
    \subfloat[Barlow Twins dropout RoBERTa]
    {\includegraphics[scale=0.44]{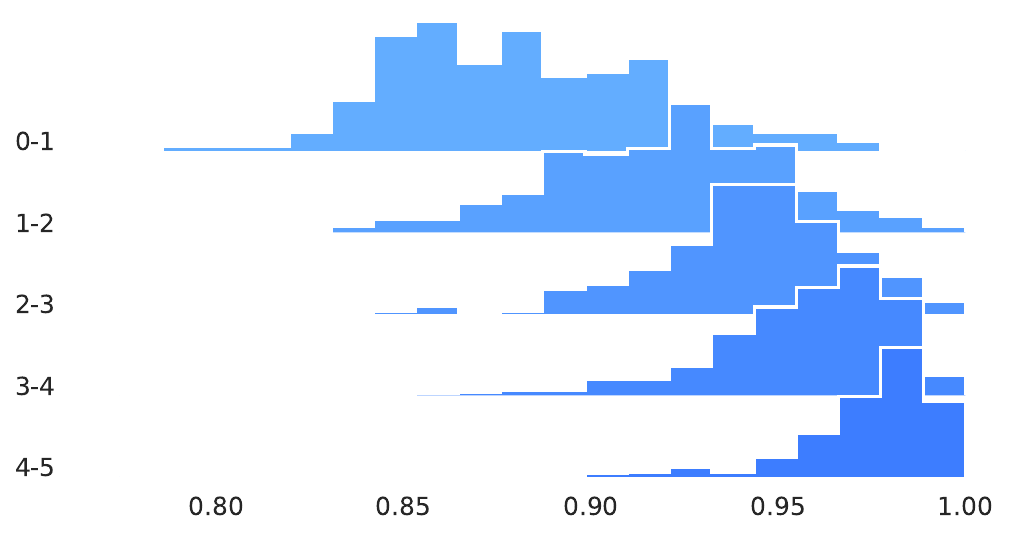}}
    
    \subfloat[VICReg shuffle RoBERTa]
    {\includegraphics[scale=0.44]{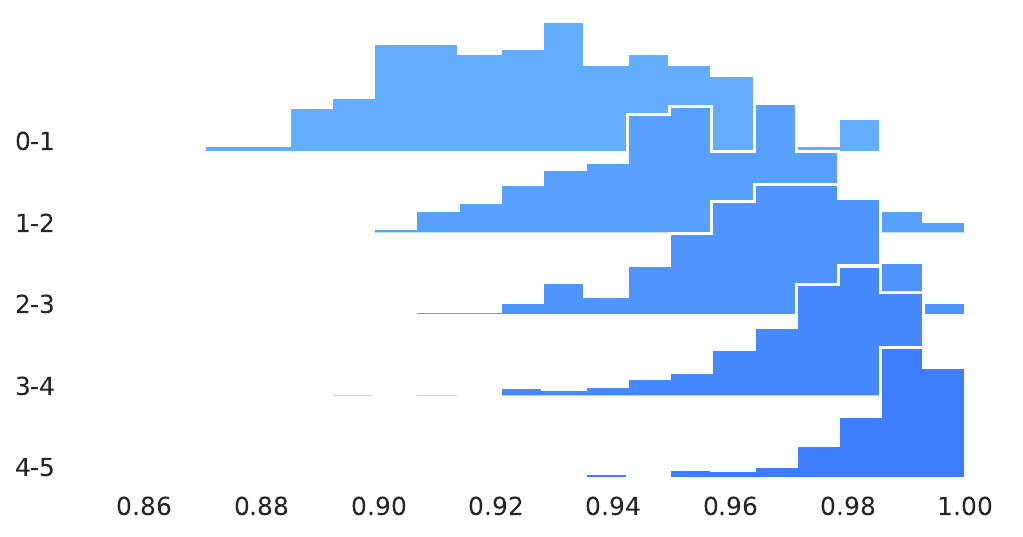}}
    \caption{Histograms of cosine similarity between pairs of sentences from the STS-B test set computed with different RoBERTa models, vertically divided in groups according to human ratings of similarity. Notice the different scale of the horizontal axis.}%
    \label{fig:histos}
\end{figure}

%% file: bigtable_mteb.tex
\scriptsize
\setlength{\tabcolsep}{2.5pt}
\begin{tabular}{l c c c c c | c c}
\hline
    Dataset & SimCSE & VICReg & VICReg & Barlow Twins & Barlow Twins & Barlow Twins & Barlow Twins \\ 
      &   & (dropout) & (shuffle) & (dropout) & (shuffle) & (NLI) & (WikiAuto) \\
    \hline
    \hline
    \multicolumn{8}{c}{Class.} \\ \hline
    AmazonCounterfactualClassification
    ~\cite{oneill2021amazoncounterfactual} & \textbf{65.5} & 64.2 & 65.2 & 65.0 & 64.1 & \underline{60.9} & 60.5 \\ 
    AmazonPolarityClassification
    ~\cite{mcauley2013amazon}
    & \textbf{76.6} & 63.3 & 64.6 & 72.9 & 62.9 & \underline{62.7} & 62.1 \\ 
    AmazonReviewsClassification
    ~\cite{mcauley2013amazon}
    & \textbf{35.0} & 29.0 & 29.8 & 33.1 & 28.7 & 28.8 & \underline{30.4} \\ 
    Banking77Classification
    ~\cite{casanueva2020banking77}
    & \textbf{78.1} & 77.3 & 76.9 & 77.9 & 76.1 & \underline{75.6} & 67.6 \\ 
    EmotionClassification
    ~\cite{saravia2018emotion}
    & \textbf{46.8} & 42.9 & 44.3 & 44.5 & 46.0 & \underline{42.7} & 40.5 \\ 
    ImdbClassification
    ~\cite{maas2011imdb}
    & \textbf{73.5} & 64.9 & 65.0 & 72.0 & 62.4 & \underline{63.0} & 57.4 \\ 
    MassiveIntentClassification
    ~\cite{fitzgerald2022massive}
    & 61.5 & 61.1 & 64.7 & \textbf{64.8} & 57.6 & \underline{60.5} & 58.8 \\ 
    MassiveScenarioClassification
    ~\cite{fitzgerald2022massive} 
    & 69.4 & 70.0 & 73.6 & \textbf{73.7} & 62.0 & \underline{70.9} & 69.5 \\ 
    MTOPDomainClassification
    ~\cite{li2020mtop} 
    & 85.1 & 85.9 & \textbf{88.1} & 88.0 & 80.9 & \underline{84.4} & 81.4 \\ 
    MTOPIntentClassification
    ~\cite{li2020mtop} 
    & 61.3 & 59.8 & 64.8 & \textbf{68.3} & 59.0 & \underline{56.0} & 51.0 \\ 
    ToxicConversationsClassification
    ~(\href{https://www.kaggle.com/competitions/jigsaw-unintended-bias-in-toxicity-classification}{url}) 
    & 68.6 & 66.4 & 66.8 & \textbf{69.9} & 64.2 & 66.3 & \underline{66.5} \\ 
    TweetSentimentExtractionClassification
    (\href{https://www.kaggle.com/competitions/tweet-sentiment-extraction}{url})
    & \textbf{54.0} & 50.4 & 51.8 & 52.4 & 48.9 & 51.3 & \underline{51.6} \\ 
    \hline
    
    \multicolumn{8}{c}{Clust.} \\ \hline
    ArxivClusteringP2P$^\diamondsuit$
    & 32.9 & 34.9 & 33.7 & \textbf{35.2} & 33.1 & \underline{38.6} & 33.5 \\ 
    ArxivClusteringS2S$^\diamondsuit$
    & 21.4 & 21.8 & \textbf{23.5} & 23.0 & 17.9 & \underline{25.8} & 23.6 \\ 
    BiorxivClusteringP2P$^\diamondsuit$
    & 30.1 & 31.5 & 30.4 & \textbf{31.7} & 30.8 & \underline{36.0} & 30.0 \\ 
    BiorxivClusteringS2S$^\diamondsuit$
    & 22.1 & 22.9 & \textbf{24.6} & 23.9 & 16.1 & \underline{26.1} & 22.0 \\ 
    MedrxivClusteringP2P$^\diamondsuit$
    & 26.9 & \textbf{29.0} & 27.4 & 28.5 & 28.8 & \underline{31.2} & 28.0 \\ 
    MedrxivClusteringS2S$^\diamondsuit$
    & 24.9 & 25.4 & \textbf{26.0} & \textbf{26.0} & 21.3 & \underline{28.3} & 25.6 \\ 
    RedditClustering
    ~\cite{geigle2021clustering}
    & 33.9 & 40.1 & 35.0 & \textbf{41.2} & 28.7 & \underline{47.0} & 41.7 \\ 
    RedditClusteringP2P$^\diamondsuit$
    & 47.2 & 48.8 & 43.1 & \textbf{50.4} & 46.3 & \underline{52.5} & 46.9 \\ 
    StackExchangeClustering
    ~\cite{geigle2021clustering}
    & 46.3 & 48.2 & 49.3 & \textbf{50.9} & 38.0 & \underline{51.9} & 49.1 \\ 
    StackExchangeClusteringP2P$^\diamondsuit$
    & 29.5 & \textbf{30.7} & 30.0 & 30.0 & 28.5 & 30.5 & \underline{33.1} \\ 
    TwentyNewsgroupsClustering
    ~(\href{https://scikit-learn.org/0.19/datasets/twenty_newsgroups.html}{url}) 
    & 23.8 & \textbf{33.5} & 33.1 & 31.9 & 19.4 & \underline{37.2} & 34.8 \\ 
    \hline
    
    \multicolumn{8}{c}{PairClass.} \\ \hline
    SprintDuplicateQuestions
    ~\cite{shah2018adversarial}
    & \textbf{86.4} & 70.7 & 77.1 & 74.1 & 88.5 & \underline{84.2} & \underline{84.2} \\ 
    TwitterSemEval2015
    ~\cite{xu2015semeval}
    & 56.8 & 56.3 & 56.3 & \textbf{59.1} & 51.8 & \underline{51.3} & 43.6 \\ 
    TwitterURLCorpus
    ~\cite{lan2017sentential}
    & \textbf{80.4} & 77.6 & 78.8 & 78.8 & 78.9 & \underline{78.3} & 75.4 \\ 
    \hline
    
    \multicolumn{8}{c}{Rerank.} \\ \hline
    AskUbuntuDupQuestions
    ~(\href{https://github.com/taolei87/askubuntu}{url})  
    & \textbf{53.3} & 51.7 & 51.9 & 52.5 & 51.9 & \underline{52.2} & 50.4 \\ 
    MindSmallReranking
    ~\cite{wu2020mind}
    & 29.4 & 29.2 & \textbf{30.3} & 29.6 & 27.9 & 30.0 & \underline{31.1} \\ 
    SciDocsRR
    ~\cite{cohan2020scidocs}
    & 66.9 & 65.5 & \textbf{68.7} & 67.5 & 62.0 & \underline{69.7} & 66.0 \\ 
    StackOverflowDupQuestions
    ~\cite{liu2018linkso}
    & \textbf{39.8} & 38.1 & 38.2 & 39.6 & 39.5 & \underline{38.4} & 34.8 \\ 
    \hline
    
    \multicolumn{8}{c}{Retr.$^\spadesuit$} \\ \hline
    ArguAna 
    & 34.7 & 43.8 & 42.6 & \textbf{43.9} & 35.6 & \underline{44.1} & 40.6 \\ 
    ClimateFEVER
    & 14.5 & 12.8 & 13.0 & \textbf{19.2} & 14.2 & 18.2 & \underline{22.0} \\ 
    CQADupstackRetrieval
    & \textbf{20.4} & 13.9 & 17 & 20.0 & 18.7 & \underline{19.4} & 18.3 \\ 
    DBPedia
    & \textbf{15.7} & 12.0 & 13.2 & 15.2 & 12.8 & \underline{17.6} & 17.2 \\ 
    FEVER
    & \textbf{28.4} & 12.6 & 15.9 & \textbf{28.4} & 17.1 & 25.2 & \underline{33.7} \\ 
    FiQA2018
    & 12.6 & 11.6 & 11.3 & \textbf{14.4} & 10.3 & \underline{16.1} & 11.3 \\ 
    HotpotQA
    & \textbf{31.4} & 16.5 & 16.8 & 25.0 & 29.7 & 26.7 & \underline{36.2} \\ 
    MSMARCO
    & \textbf{8.8} & 5.4 & 6.1 & 7.8 & 7.8 & 8.6 & \underline{12.6} \\
    NFCorpus
    & \textbf{14.3} & 9.1 & 10.6 & 11.7 & 10.1 & 15.6 & \underline{18.7} \\
    NQ
    & 12.3 & 7.3 & 8.9 & \textbf{13.6} & 9.0 & 12.3 & \underline{15.4} \\ 
    QuoraRetrieval
    & \textbf{80.4} & 78.5 & 79.5 & 79.6 & 78.3 & \underline{78.2} & 75.0 \\ 
    SCIDOCS
    & 6.9 & 5.7 & 6.6 & \textbf{7.4} & 7.2 & \underline{10.5} & 9.5 \\ 
    SciFact
    & \textbf{34.1} & 27.3 & 24.3 & 25.6 & 34.7 & \underline{35.2} & 34.5 \\ 
    Touche2020
    & 10.9 & 10.4 & 9.7 & \textbf{11.9} & 10.6 & \underline{13.1} & 10.5 \\ 
    TRECCOVID
    & 28 & 30.9 & 35.1 & \textbf{38.0} & 26.1 & \underline{36.7} & 33.6 \\ \hline
    
    \multicolumn{8}{c}{STS} \\ \hline
    BIOSSES~(\href{https://tabilab.cmpe.boun.edu.tr/BIOSSES/DataSet.html}{url})
    & 67.7 & 51.1 & 56.9 & 56.9 & \textbf{69.5} & 58.8 & \underline{68.6} \\ 
    SICK-R~\cite{agirre2014semeval}
    & 68.9 & 67.9 & 70.1 & \textbf{70.6} & 64.8 & 64.3 & \underline{67.4} \\ 
    STS12$^\heartsuit$
    & \textbf{70.2} & 64.2 & 63.2 & 62.5 & 65.4 & \underline{66.5} & 66.3 \\ 
    STS13$^\heartsuit$
    & \textbf{81.8} & 78.7 & 77.3 & 77.6 & 77.7 & \underline{77.3} & 77.2 \\ 
    STS14$^\heartsuit$
    & \textbf{73.2} & 68.1 & 66.6 & 68.1 & 70.5 & \underline{67.7} & 67.4 \\ 
    STS15$^\heartsuit$
    & \textbf{81.4} & 78.5 & 76.3 & 76.2 & 80.4 & \underline{75.3} & 74.1 \\ 
    STS16$^\heartsuit$
    & \textbf{80.7} & 77.5 & 77.4 & 79.3 & 76.0 & \underline{75.0} & 74.7 \\
    STS17$^\heartsuit$
    & 81.8 & 81.2 & 81.6 & \textbf{82.0} & 80.8 & 78.2 & \underline{79.8} \\
    STS22$^\heartsuit$
    & 57.7 & 60.2 & 59.8 & \textbf{61.0} & 60.8 & \underline{61.9} & 55.5 \\
    STSBenchmark$^\heartsuit$
    & \textbf{80.1} & 78.0 & 76.9 & 76.6 & 75.6 & 75.1 & \underline{75.2} \\ \hline
    
    \multicolumn{8}{c}{Summ.} \\ \hline
    SummEval~
    \cite{fabbri2020summeval} 
    & 27.6 & 28.7 & \textbf{29.2} & 28.9 & 27.6 & 27.5 & \underline{31.1} \\ 
    \hline

\end{tabular}